\documentclass[journal]{IEEEtran}

\usepackage{graphicx} 

\usepackage{color} 
\usepackage{subcaption}
\usepackage{amsfonts}
\usepackage{amsmath}
\usepackage{bbm}
 
\newcommand{\myrot}[1]{\rotatebox{90}{#1}}

\usepackage{subcaption}

\begin{document}

\title{Saliency Guided Self-attention Network for Weakly and Semi-supervised Semantic Segmentation}


\author{
	Qi Yao, Xiaojin Gong
	\\
	Zhejiang University, Hangzhou, Zhejiang, 310027, China
	\\
	\{yaoqi\_isee, gongxj\}@zju.edu.cn	
}

\maketitle

\begin{abstract}
Weakly supervised semantic segmentation (WSSS) using only image-level labels can greatly reduce the annotation cost and therefore has attracted considerable research interest. However, its performance is still inferior to the fully supervised counterparts. To mitigate the performance gap, we propose a saliency guided self-attention network (SGAN) to address the WSSS problem. The introduced self-attention mechanism is able to capture rich and extensive contextual information but may mis-spread attentions to unexpected regions. In order to enable this mechanism to work effectively under weak supervision, we integrate class-agnostic saliency priors into the self-attention mechanism and utilize class-specific attention cues as an additional supervision for SGAN. Our SGAN is able to produce dense and accurate localization cues so that the segmentation performance is boosted. Moreover, by simply replacing the additional supervisions with partially labeled ground-truth, SGAN works effectively for semi-supervised semantic segmentation as well. Experiments on the PASCAL VOC 2012 and COCO datasets show that our approach outperforms all other state-of-the-art methods in both weakly and semi-supervised settings.
\end{abstract}

\section{Introduction}
\label{sec:introduction}
Semantic segmentation aims to predict a semantic label for each pixel in an image. Based upon the fundamental Fully Convolutional Networks (FCNs)~\cite{Long2015fcn}, various techniques such as dilated convolution~\cite{chen2017deeplab}, spatial pyramid pooling~\cite{Yang2018denseaspp}, and encoder-decoders~\cite{Fischer2015unet} have been developed in the last decade. These techniques gradually improve segmentation accuracy via exploiting extensive contextual information. Recently, the self-attention mechanism~\cite{yuan2018ocnet,fu2019dual,Huang2019criss-cross} has been successfully employed to capture richer contextual information and boost the segmentation performance further. Although the above-mentioned methods have achieved high performance in semantic segmentation, they all work under full supervision. This supervision manner requires a large amount of pixel-wise annotations for training, which are very expensive and time-consuming. 

To reduce the annotation burden, different supervision forms such as bounding boxes~\cite{Xu2015weak}, scribbles~\cite{lin2016scribble}, and image-level tags~\cite{kolesnikov2016seed} have been considered for semantic segmentation. Among them, the form of using image-level tags has attracted major attention because of its minimal annotation cost as well as its great challenge. Recent work~\cite{zhou2016learning} has shown that convolutional neural networks (CNNs) have the localization ability even if only image-level tags are used. This observation has inspired many weakly-supervised semantic segmentation (WSSS) researches. However, attentions in the class activation maps (CAMs)~\cite{zhou2016learning} inferred from image classification networks tend to focus on small discriminative parts of objects. The object location cues (also referred to as \textit{\textbf{seeds}}) retrieved from these CAMs are too sparse to effectively train a segmentation model. Therefore, a great amount of effort has been devoted to recover dense and reliable seeds~\cite{wei2017object,huang2018weakly,wei2018revisiting,lee2019ficklenet,Jing2019dilated}. 

\begin{figure}[t]
	\centering
	\includegraphics[width=.45\textwidth]{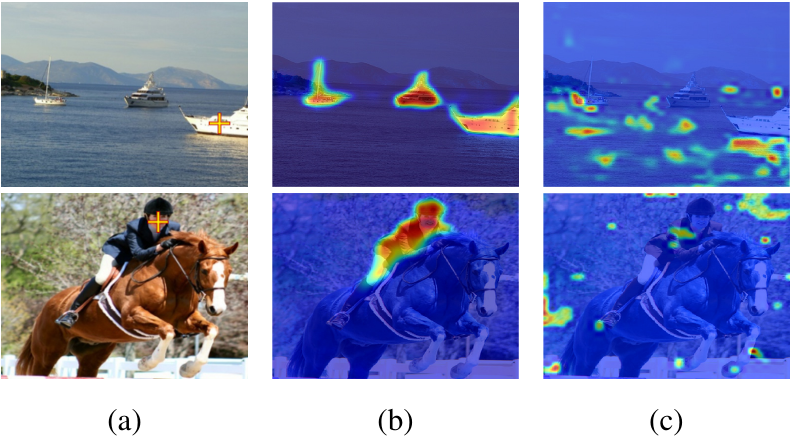}
	\caption{Illustrations of context attention maps learned by the self-attention scheme. (a) shows images where pixels of interest are marked by '+'. (b) presents the attention maps learned in a fully-supervised segmentation network, in which the pixels belonging to the same category with the selected pixels are highlighted. (c) shows the results learned in a weakly-supervised scenario, in which the information of the selected pixels is mis-spread to unexpected regions.}
	\label{fig:self-attention}
\end{figure}

In this paper, we aim to take advantage of the self-attention mechanism to mine high-quality seeds. As validated in~\cite{fu2019dual,yuan2018ocnet}, this mechanism is able to successfully capture long-range contextual dependencies in fully-supervised semantic segmentation. However, it encounters the following challenges when applied to WSSS. (1) Some foreground objects may always be co-occurrent with the same background, like 'boat' and 'water', leading to a pathological bias~\cite{li2018tell}; (2) The global average pooling (GAP), which is commonly used in classification networks to aggregate pixel-wise responses into image-level label scores, encourages all responses to be high; (3) In the self-attention scheme, each pixel directly contributes to all other pixels and vice versa. These factors may result in a mis-spread of attentions from discriminative parts to unexpected regions. Typical examples are illustrated in Figure~\ref{fig:self-attention}. In the fully-supervised setting, the information of the selected discriminative pixels can be correctly propagated to the pixels belonging to the same category. Contrastively, the discriminative information is diffused to the regions of the background and other categories under weak supervision.

To address the above-mentioned problems and enable the self-attention mechanism to capture long-range contextual information correctly under weak supervision, we construct a self-attention network that leverages the class-agnostic saliency as a guidance. A saliency map provides a rough detection of foreground objects so that it can prevent attentions from spreading to unexpected background regions. To further reduce information diffusion among foreground categories, we integrate the class-specific attention cues as additional supervision. The integration of these prior cues is implemented in our network via a joint learning of a seed segmentation branch and an image classification branch. By all these means, our network generates high quality seeds so that the segmentation performance is boosted. 

Our work distinguishes itself from the others as follows:
\begin{itemize}
	\item We propose a saliency-guided self-attention network (SGAN) for weakly supervised semantic segmentation. It integrates class-agnostic saliency maps and class-specific attention cues to enable the self-attention mechanism to work effectively under weak supervision. Moreover, these two types of priors are fused adaptively in our SGAN to help the generation of high quality seeds. 
	\item In our network, both the seed segmentation branch and the image classification branch can produce high quality seed results. The ensemble of two results improves the quality of seeds further. 

	\item By simply replacing saliency maps and attention cues with partially labeled segmentation ground-truth, SGAN can work effectively for semi-supervised semantic segmentation as well. 

	\item Our approach achieves state-of-the-art performance on the PASCAL VOC 2012 and COCO datasets in both weakly and semi-supervised settings.
\end{itemize}

\section{Related Work}

\subsection{Weakly-supervised Semantic Segmentation}
Various supervision forms have been exploited for weakly-supervised semantic segmentation (WSSS). Here, we focus on the works using image-level tags. Most recent methods solve the WSSS problem by first mining reliable seeds and then take them as proxy ground-truth to train segmentation models. Thus, a great amount of effort has been devoted to generate high-quality seeds.

A group of approaches take the class activation maps (CAMs)~\cite{zhou2016learning} generated from classification networks as initial seeds. Since CAMs only focus on small discriminative regions which are too sparse to effectively supervise a segmentation model, various techniques such as adversarial erasing~\cite{wei2017object,li2018tell,hou2018self,chaudhry2017discovering}, attention accumulation~\cite{jiang2019integral}, and region growing~\cite{huang2018weakly,Sun2019saliency,Shimoda_2019_ICCV} have been developed to expand sparse object seeds. Another research line introduces dilated convolutions of different rates~\cite{wei2018revisiting,Jing2019dilated,lee2019ficklenet,fan2018cian} to enlarge receptive fields in classification networks and aggregate multiple attention maps to achieve dense localization cues. In this work, we adopt the self-attention mechanism to capture richer and more extensive contextual information in order to mine high quality seeds.

\subsection{Self-attention Mechanism}
The self-attention mechanism~\cite{lin2017structured} computes the context at each position as a weighted sum of all positions. Its superiority in capturing long-range dependencies has been recently validated by various vision tasks~\cite{Wang2018nonlocal,hu2018relation,yuan2018ocnet,fu2019dual}. Particularly, in semantic segmentation, Yuan and Wang~\cite{yuan2018ocnet} integrated this mechanism into pyramid structures to capture multi-scale contextual information; Fu \textit{et al.}~\cite{fu2019dual} constructed a dual attention network to capture dependencies in both spatial and channel dimensions; Huang \textit{et al.}~\cite{Huang2019self} proposed an interlaced sparse approach to improve the efficiency of the self-attention mechanism; and Huang \textit{et al.}~\cite{Huang2019criss-cross} designed a recurrent criss-cross attention module to efficiently harvest the contextual information. These methods significantly boost the segmentation performance, but all of them perform under full supervision. Although Sun and Li~\cite{Sun2019saliency} utilized the self-attention scheme for WSSS, they only used this scheme to learn a saliency detector that is trained also in a fully-supervised manner. In our work, we apply the self-attention scheme to a weakly-supervised scenario which is more challenging.

\subsection{Saliency Guidance for WSSS}
Salient object detection (SOD)~\cite{xiao2018deep} produces class-agnostic saliency maps that distinguish foreground objects from the background. The results of SOD have been extensively used in weakly-supervised semantic segmentation. For instance, many methods~\cite{wei2017object,huang2018weakly,lee2019ficklenet,fan2018cian,li2018tell,wei2018revisiting} exploited saliency maps to generate background seeds. Moreover, Wei \textit{et al.}~\cite{wei2017stc} adopted a self-paced learning strategy to learn a segmentation model that was initialized under the full supervision of saliency maps of simple images. Sun and Li~\cite{Sun2019saliency} utilized saliency maps to guide a CAM-seeded region growing process to expand object regions. Fan \textit{et al.}~\cite{fan2018associating} used instance-level saliency maps to construct and partition similarity graphs for WSSS. In addition, Chaudhry \textit{et al.}~\cite{chaudhry2017discovering}, Oh \textit{et al.}~\cite{oh2017exploiting}, and Wang \textit{et al.}~\cite{wang2018weakly} combined class-agnostic saliency maps with class-specific attention cues like us to obtain dense seed. But their combinations are implemented in user-defined ways. In contrast, our saliency maps and attention cues are adaptively fused within the proposed self-attention network. 

\begin{figure*}[tbp]
	\centering
	\includegraphics[width=\linewidth]{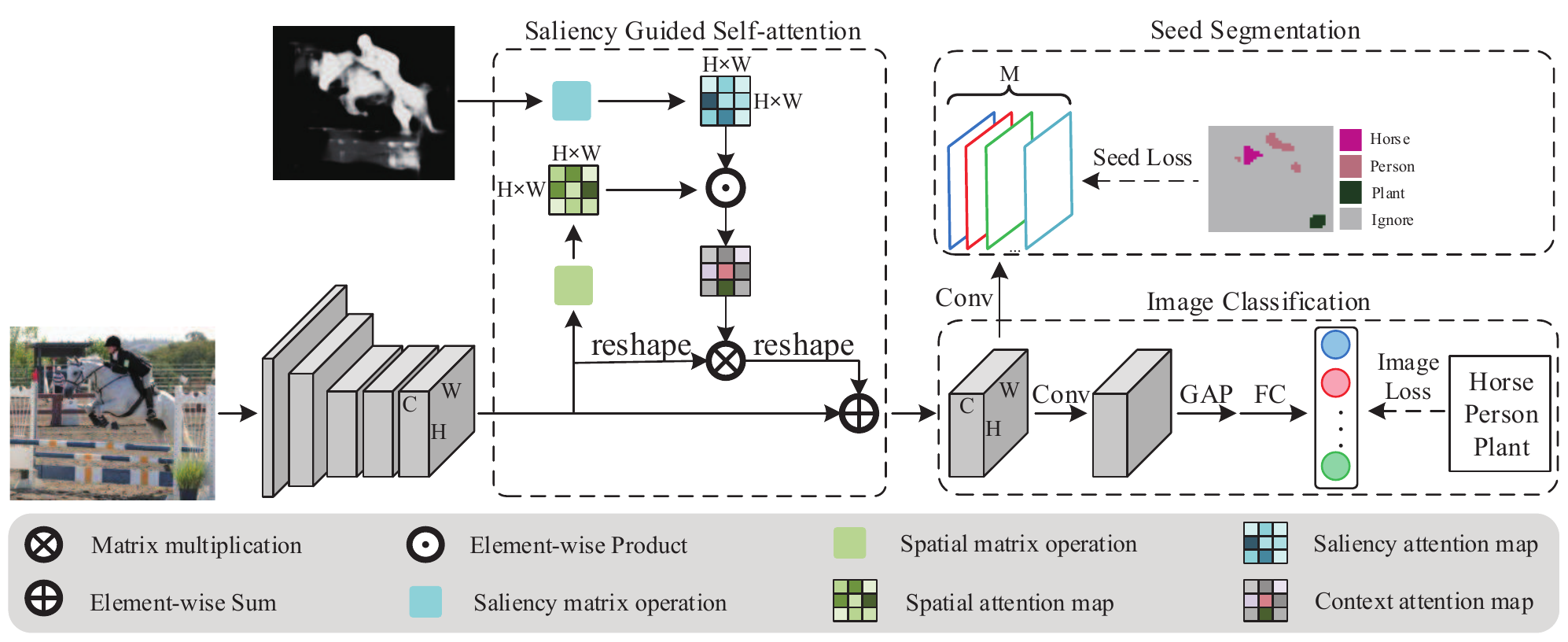}
	\caption{An overview of the proposed saliency guided self-attention network.}
	\label{fig:architecture}
\end{figure*}


\section{The Proposed Approach}
The proposed approach for weakly and semi-supervised semantic segmentation is divided into two parts: (1) learning a saliency guided self-attention network to generate dense and accurate seeds, and (2) utilizing the high-quality seeds as proxy ground-truth to train a semantic segmentation model. The details are introduced in the followings.

\subsection{Saliency Guided Self-attention Network}

\subsubsection{Network Architecture} 
The overview of our proposed saliency guided self-attention network (SGAN) is illustrated in Figure~\ref{fig:architecture}. It consists of three components: (1) a CNN backbone to learn deep feature representations; (2) a saliency guided self-attention module that propagates attentions from small discriminative parts to non-discriminative regions via capturing long-range contextual dependencies; (3) an image classification branch together with a seed segmentation branch to supervise the training of the entire network. 

We adopt a slightly modified VGG-16~\cite{kolesnikov2016seed} network as the backbone for feature extraction. The last two pooling layers are removed in order to increase the resolution of the output feature maps. Note that, unlike previous works~\cite{wei2018revisiting,fan2018cian,lee2019ficklenet} that enlarge the dilation rate of convolution kernels in conv5 block, we avoid the usage of dilated convolution and instead use the self-attention module to capture more extensive contexts.

\subsubsection{Saliency Guided Self-attention Module}
\label{SGAN} 
This module aims to take advantage of the self-attention mechanism to capture rich contextual information that is essential for discovering integral extent of objects and retrieving high-quality seeds. The self-attention mechanism has demonstrated its effectiveness in capturing long-range dependencies under full supervision~\cite{fu2019dual,yuan2018ocnet}. However, simply integrating it into a weakly-supervised network may suffer from a severe mis-spread problem as introduced in Section~\ref{sec:introduction}. Thus, we propose to incorporate class-agnostic saliency priors to prohibit the spread of attentions from discriminative object regions to the background. 

We formally describe the saliency guided self-attention module as follows. This module takes the feature map output from the VGG's conv5 block, which is denoted as $\mathbf{X} \in \mathbb{R}^{C \times H \times W}$, together with a saliency map as the inputs. With the input feature map, a sequence of spatial matrix operations are performed to generate a spatial attention map $\mathbf{P} \in \mathbb{R}^{N \times N}$, where $N = H \times W$ is the number of positions. More specifically, $\mathbf{X}$ is first fed into two $1 \times 1$ convolutions for linear embedding and generating a key feature map $\mathbf{K}\in \mathbb{R}^{C\times H\times W}$ and a query feature map $\mathbf{Q}\in \mathbb{R}^{C\times H\times W}$ respectively. These two feature maps are further reshaped to $\mathbb{R}^{C \times N}$. Then, the \textit{\textbf{spatial attention map}} $\mathbf{P}$ is generated by computing the inner product of channel-wise features from any two positions of $\mathbf{K}$ and $\mathbf{Q}$. That is, 
\begin{equation}
P_{ij} = K_i^T Q_j,
\end{equation}
where $\{i, j\} \in \{1, 2, ..., N\}$ are the indexes of positions, and $\{K_i, Q_j\} \in \mathbb{R}^{C\times 1}$ are the channel-wise features. $P_{ij}$ measures the similarity of the features extracted at position $i$ and $j$. Note that different pairwise functions~\cite{Wang2018nonlocal} can be used for the similarity measurement, we take the inner product because it is simple but effective enough. 


For the input saliency map, we first threshold it to get a binary mask $\mathbf{B}$ and reshape it to $\mathbb{R}^{N \times 1}$. After that, a \textit{\textbf{saliency attention map}} $\mathbf{S}\in \mathbb{R}^{N\times N}$ is computed by
\begin{equation}
S_{ij} = \mathbbm{1}(B_i == B_j),
\end{equation}
where $\mathbbm{1}$ is an indicator function. It equals one if both positions $i$ and $j$ are salient or non-salient.

Then, the \textit{\textbf{context attention map}} $\mathbf{D} \in \mathbb{R}^{N\times N}$ is generated via an element-wise production between the spatial attention map $\mathbf{P}$ and the saliency attention map $\mathbf{S}$, followed by a linear normalization:
\begin{equation}
D_{ij} = \frac{P_{ij} \cdot S_{ij}}{\sum_{j=1}^{N} P_{ij} \cdot S_{ij}}.
\end{equation}

Once the context attention map $\mathbf{D}$ is obtained, we use it to enhance the original feature map $\mathbf{X}$. Specifically, we reshape $\mathbf{X}$ to $\mathbb{R}^{C\times N}$ and conduct a matrix multiplication between $\mathbf{X}$ and the transpose of $\mathbf{D}$. Then we reshape the result back to $\mathbb{R}^{C\times H\times W}$ and perform an element-wise summation with $\mathbf{X}$ to obtain the enhanced features $\mathbf{E}\in \mathbb{R}^{C\times H\times W}$. That is,
\begin{equation} \label{eq:agg}
E_i = \gamma \sum_{j = 1}^{N} (D_{ij} \cdot X_j) + X_i,
\end{equation}
where $\gamma$ is a parameter initialized as 0~\cite{fu2019dual} and gradually learned in training. Equation~(\ref{eq:agg}) indicates that each position of $\mathbf{E}$ is the sum of similarity-weighted features at all positions and the original features. Therefore, this module captures contextual information from a whole image. By this means, attentions on small discriminative parts of objects can be propagated to non-discriminative object regions, but not to the background because of the guidance of saliency.



%

\subsubsection{Integrating Class-specific Attention Cues} The class-agnostic saliency maps introduced above can only roughly separate foreground objects from the background, but provide no information about semantic categories. In order to prevent our SGAN from mis-spreading attentions among objects of different categories, we propose to integrate the class-specific attention cues obtained by the CAM method~\cite{zhou2016learning} from a classification network as additional supervision. 


Specifically, we construct a segmentation branch in our SGAN. It takes the enhanced feature $\mathbf{E}$ as the input and goes through a convolutional layer to produce $M$ segmentation maps, each of which corresponds to a foreground category. Meanwhile, we retrieve reliable but sparse foreground object seeds by thresholding the class activation maps obtained from the VGG-16 classification network with a high confidence value (empirically set to 0.3 in this work) and use them to supervise the segmentation maps. The corresponding seed loss $L_{seed}$ is defined by
\begin{equation}
L_{seed} = - \frac{1}{\sum_{z \in Z} |\Lambda_z|} \sum_{z \in Z} \sum_{u \in \Lambda_z} log \Phi_{z, u}.
\end{equation} 
Here, $Z$ denotes the set of foreground classes that present in an image and $\Lambda_z$ is a set of seed locations corresponding to class $z$. $|\cdot|$ is the cardinality of the set. $\Phi_{z, u}$ denotes the probability of class $z$ at position $u$ in the segmentation maps. Note that, in contrast to the seeding loss defined in~\cite{kolesnikov2016seed,huang2018weakly} that considers both foreground and background categories, our loss only takes into account the foreground classes. 


\subsubsection{Training SGAN} 
The network also has an image classification branch that is supervised by image-level labels. Let us denote the classification probability as $\tau \in \mathbb{R}^{M\times 1}$ and the corresponding image-level label as $y=[y_1,...,y_M] \in \{1, -1 \}$, which indicates the presence or absence of foreground object categories. Then the classification loss $L_{cls}$ is defined by the sigmoid cross entropy. That is,
\begin{equation}
L_{cls}=-\frac{1}{M} \sum_{m=1}^{M}log(y_m \cdot (\tau_m - \frac{1}{2}) + \frac{1}{2}).
\end{equation}

The overall loss for training our saliency guided self-attention network is defined by
\begin{equation}
\label{eq:sganloss}
L = L_{cls} + \lambda L_{seed},
\end{equation}
where $\lambda$ is a weighting factor to balance the two terms.

\subsubsection{Generating High-quality Seeds} 
Once the proposed SGAN is trained, we note that there are two possible ways to get class activation maps (CAMs). One is following the common practice~\cite{zhou2016learning} to infer the CAMs from the image classification branch. The other is directly taking the segmentation maps output from the seed segmentation branch. Either may be used to retrieve dense and accurate seeds. But we find out that the combination of them improves the seeds' quality further because they are complementary in some scenarios as will be shown in Section~\ref{sec:ablation}. Therefore, we take an element-wise summation of these two results to generate the final class activation maps. 

Then, for each foreground class, we retrieve object seeds by thresholding the corresponding class activation map with a high value $\alpha$. In addition, we retrieve background seeds by thresholding the input saliency map with a low value $\beta$. Following~\cite{kolesnikov2016seed,huang2018weakly,wei2018revisiting}, we set $\alpha=0.2$ and $\beta=0.06$ in our experiments.

\subsection{Training the Segmentation Network} 
After obtaining the high-quality seeds, we can use them as proxy ground-truth labels to train an arbitrary semantic segmentation network. In this work, we adopt the balanced seed loss $L_{balance\_seed}$ proposed in DSRG~\cite{huang2018weakly} for the seed supervision. It is
\begin{equation}
\begin{split}
L_{balance\_seed} = -\frac{1}{\sum_{z \in Z} |\Lambda_z|} \sum_{z \in Z} \sum_{u \in \Lambda_z} log \Phi_{z,u} \\
-\frac{1}{\sum_{z \in \overline{Z}} |\Lambda_z|} \sum_{z \in \overline{Z}} \sum_{u \in \Lambda_z} log \Phi_{z,u},
\end{split}
\end{equation}
where $\overline{Z}$ denotes the background. $Z$, $\Lambda_z$, and $\Phi_{z, u}$ holds the same definitions as previous. 

We further exploit the boundary constraint loss used in both DSRG~\cite{huang2018weakly} and SEC~\cite{kolesnikov2016seed} to encourage segmentation results to match up with object boundaries. Let us denote $\mathbf{I}$ as the input image and $\mathbf{R}(\mathbf{I}, \Phi)$ as the output probability map of the fully-connected CRF. Then the boundary constraint loss is defined as the mean KL-divergence between the segmentation map and the output of CRF:
\begin{equation}
L_{boundary}=\frac{1}{N} \sum_{u=1}^{N} \sum_{z=1}^{M+1} \mathbf{R}_{z,u}(\mathbf{I}, \Phi) log \frac{\mathbf{R}_{z,u}(\mathbf{I}, \Phi)}{\Phi_{z, u}}.
\end{equation}
Thus, the total loss for training the segmentation model is $L = L_{balance\_seed} + L_{boundary}$. 

\subsection{SGAN Under Semi-supervision}
\label{sec:semi-sgan}
In the semi-supervised setting, a small number of training images are provided with strong pixel-level labels and the rest have image-level tags only. For these strongly annotated images, we replace their saliency maps with the binary foreground masks derived from the ground-truth annotations, and meanwhile use the ground-truth to supervise the seed segmentation branch. These simple replacements help the proposed SGAN to learn a better model and generate higher-quality seeds for all images. 

For the training of the segmentation network, we also have the ground-truth to take place of the generated seeds for these strongly annotated images. Except this, we keep the training loss under semi-supervision the same as that in the weakly-supervised setting.

\section{Experiments}
\subsection{Experimental Setup}

\subsubsection{Dataset and Evaluation Metric} The proposed approach is evaluated on the PASCAL VOC 2012 segmentation benchmark~\cite{everingham2010pascal} and the COCO~\cite{lin2014microsoft} dataset.

\textbf{PASCAL VOC:} This dataset provides pixel-wise annotations for 20 object classes and one background class. It contains 1464 images for training, 1449 images for validation and 1456 images for testing. Following the common practice~\cite{kolesnikov2016seed,huang2018weakly,wei2018revisiting}, we augment the training set to 10,582 images. Our network is trained on the augmented training set only using image-level annotations and evaluated on the validation set in terms of the mean intersection-over-union (mIoU) criterion. Evaluation results of the test set are obtained by submitting our prediction results to the official PASCAL VOC evaluation server. 

\textbf{COCO:} This dataset contains 80k images for training and 40k images for validation. Our network is trained on the training set with only image-level annotations and evaluated on the validation set in terms of mIoU for 81 categories.

\subsubsection{Training Details}
The saliency guided self-attention network is built on the VGG-16 network pre-trained on ImageNet. The remaining parameters of our SGAN are randomly initialized. Following~\cite{wei2018revisiting}, we use S-Net~\cite{xiao2018deep} to get a class-agnostic saliency map for each image. SGD with mini-batch is used for training. The batch size is set to 15, the momentum is 0.9 and the weight decay is 0.0005. Input images are resized to $321\times 321$ and no data augmentation except randomly horizontal flip are adopted. We train the SGAN for 8,000 iterations. The initial learning rate is 0.001 and it is decreased by a factor of 10 every 2,000 iterations.

The semantic segmentation model is chosen to use the Deeplab-ASPP~\cite{chen2017deeplab} network in order to compare with other WSSS works. Both VGG-16 and ResNet-101 backbones are investigated. The batch size is set to 15, the momentum is 0.9 and the weight decay is 0.0005. Input images are resized to $353\times 353$ and randomly cropped to $321\times 321$ for training. Horizontal flip and color jittering are employed for data augmentation. We train the segmentation model for 12,000 iterations. The initial learning rate is 0.001 and it is decreased by a factor of 0.33 every 2,000 iterations. 

\subsubsection{Reproducibility} We implement our SGAN on PyTorch~\cite{paszke2017automatic} for training and producing high-quality seeds. We use the official Deeplab-ASPP code implemented on Caffe~\cite{jia2014caffe} for semantic segmentation. All the experiments are conducted on a GTX 1080Ti GPU. The code is available at https://github.com/yaoqi-zd/SGAN.

\subsection{Comparison to the State of the Art}
\subsubsection{Weakly-supervised Semantic Segmentation}\

\textbf{PASCAL VOC:} We compare our approach with other state-of-the-art WSSS methods that are also supervised only by image-level labels. For fair comparison, we separate the methods into two groups according to the backbones upon which their segmentation models are built, as listed in Table~\ref{tab:comp}. Most of existing methods use saliency maps to retrieve background seeds or even foreground seeds. Therefore, Table~\ref{tab:comp} also marks out if a method uses saliency maps. 

\begin{table}[tbhp]
	\begin{minipage}{1\linewidth}
		\centering
		\caption{Comparison of weakly-supervised semantic segmentation methods on PASCAL VOC 2012 validation and test sets in terms of mIoU (\%). For the methods using the ResNet backbone for segmentation, most of them use ResNet-101, except these marked with $^{\dagger}$ that use ResNet-38.}
		\label{tab:comp}
		\begin{tabular}{lcccc}
			\hline 
			Methods                             &Publication  & Extra guidance  &Val   &Test \\ \hline 
			\multicolumn{4}{l}{\textbf{Backbone: VGG-16 network}} \\ 
			SEC~\cite{kolesnikov2016seed}       &ECCV'16     & -                  &50.7  &51.1 \\
			AF-SS~\cite{qi2016augmented}    	&ECCV'16     & -                  &52.6  &52.7 \\
			CBTS~\cite{roy2017combining}    	&CVPR'17     & -                  &52.8  &53.7 \\
			AE\_PSL~\cite{wei2017object}    	&CVPR'17     & saliency       &55.0  &55.7 \\
			DCSP~\cite{chaudhry2017discovering} &CVPR'17	  & saliency       &58.6  &59.2 \\
			GAIN~\cite{li2018tell} 				&CVPR'18	  & saliency 				   &55.3  &56.8 \\ 
			MCOF~\cite{wang2018weakly} 			&CVPR'18	  & saliency 	   &56.2  &57.6 \\
			AffinityNet~\cite{ahn2018learning}  &CVPR'18	  & - 				   &58.4  &60.5 \\
			DSRG~\cite{huang2018weakly} 		&CVPR'18	  & saliency 	   &59.0  &60.4 \\
			MDC~\cite{wei2018revisiting} 		&CVPR'18	  & saliency 	   &60.4  &60.8 \\
			SeeNet~\cite{hou2018self} 			&NeurIPS'18  & saliency      &61.1  &60.7 \\
			AISI~\cite{fan2018associating} 		&ECCV'18	  & instance saliency  &61.3  &62.1 \\
			SGDN~\cite{Sun2019saliency} 		&PRL'19	  & saliency  	   &50.5  &51.3 \\
			DSNA~\cite{Zhang2019attention} 		&TMM'19 	  & -    &55.4  &56.4 \\
			FickleNet~\cite{lee2019ficklenet}   &CVPR'19	  & saliency  	   &61.2  &61.8 \\ 
			SSNet~\cite{Zeng_2019_ICCV} 			&ICCV'19	  & saliency    &57.1  &58.6 \\
			OAA~\cite{jiang2019integral} 		&ICCV'19	  & saliency  	   &63.1  &62.8 \\
			RRM~\cite{Zhang2020}						&AAAI'20		& -	& 60.7 & 61.0 \\
			SGAN(Ours) &- & saliency   &\textbf{64.2} &\textbf{65.0} \footnote{http://host.robots.ox.ac.uk:8080/anonymous/GLCTVA.html} \\ \hline
			
			\multicolumn{4}{l}{\textbf{Backbone: ResNet network}} \\
			DCSP~\cite{chaudhry2017discovering} &CVPR'17	 & saliency  	 &60.8  &61.8 \\
			MCOF~\cite{wang2018weakly} 			&CVPR'18	 & saliency  	 &60.3  &61.2 \\
			AffinityNet$^{\dagger}$~\cite{ahn2018learning} &CVPR'18	 &- 			 	 &61.7  &63.7 \\  
			DSRG~\cite{huang2018weakly} 		&CVPR'18	 & saliency  	 &61.4  &63.2 \\
			SeeNet~\cite{hou2018self} 			&NeurIPS'18 & saliency  	 &63.1 	&62.8 \\
			AISI~\cite{fan2018associating} 		&ECCV'18 	 & instance saliency &63.6  &64.5 \\
			CIAN~\cite{fan2018cian} 			&arXiv'18   & saliency   	 &64.1  &64.7 \\
			DFPN~\cite{Jing2019dilated} 		&TIP'19	 & -     &61.9  &62.8 \\
			DSNA~\cite{Zhang2019attention}      &TMM'19     & -  	 &58.2  &60.1 \\
			FickleNet~\cite{lee2019ficklenet} 	&CVPR'19    & saliency  	 &64.9 	&65.3 \\
			SSDD$^{\dagger}$~\cite{Shimoda_2019_ICCV} 		&ICCV'19	 &- 			     &64.9  &65.5 \\
			OAA~\cite{jiang2019integral} 		&ICCV'19 	 & saliency    &65.2  &66.4 \\
			SSENet$^{\dagger}$~\cite{SSENet}						&arXiv'19	 & - &63.3 &64.9 \\
			RRM~\cite{Zhang2020}						&AAAI'20		& -	&	66.3 & 66.5 \\
			SGAN(Ours) &- & saliency  &\textbf{67.1} &\textbf{67.2} \footnote{http://host.robots.ox.ac.uk:8080/anonymous/SINTUJ.html} \\ \hline 
		\end{tabular}
	\end{minipage}
\end{table}

Table~\ref{tab:comp} shows that our method outperforms all the previous methods on both VGG-16 and ResNet-101 backbones. In particular, AE\_PSL~\cite{wei2017object}, GAIN~\cite{li2018tell}, and SeeNet~\cite{hou2018self} use erasing techniques to get dense localization cues, which tend to identify true negative regions. AffinityNet~\cite{ahn2018learning}, DSRG~\cite{huang2018weakly}, SGDN~\cite{Sun2019saliency}, and SSDD~\cite{Shimoda_2019_ICCV} adopt region growing techniques to expand seeds. It may be hard for them to expand to non-discriminative regions if initial seeds are concentrated on extremely small discriminative parts. OAA~\cite{jiang2019integral} accumulates attention maps during the training procedure which may introduce unexpected attention regions at the early stage when the classifier is weak. MDC~\cite{wei2018revisiting}, DFPN~\cite{Jing2019dilated}, and FickleNet~\cite{lee2019ficklenet} use dilated convolutions to retrieve dense seeds, whose receptive fields are not adaptive to image contents and may result in over-expansion. In contrast, our method can achieve dense and accurate seeds, which is benefitted from the self-attention mechanism as well as the design of our SGAN network. 

It needs to be mentioned that our approach outperforms DSNA~\cite{Zhang2019attention}, which uses a spatial attention mechanism, by a great margin. Our approach also performs better than AISI~\cite{fan2018associating} that leverages strong instance-level saliency information, CIAN~\cite{fan2018cian} that utilizes cross-image affinities, and SSNet~\cite{Zeng_2019_ICCV} that jointly learn saliency and segmentation with additional pixel-wise saliency supervision. 

\textbf{COCO:} To further validate the effectiveness of our approach, we conduct experiments on the COCO dataset which is much more challenging than PASCAL VOC. Most existing methods haven't done the experiments on COCO yet, except DSRG~\cite{huang2018weakly}. Therefore, we compare our results to DSRG in Table~\ref{tab:comp_coco}, in which the results of SEC are also quoted from DSRG's paper. Table~\ref{tab:comp_coco} shows that our VGG16-based SGAN surpasses both SEC and DSRG by a large margin. In particular, our method performs excellently on those categories with large scale, such as airplane, bus and train etc, but has difficulties in handling small things that are likely to be indistinguishable with the clustered background, such as baseball glove and spoon etc.

\begin{table}[htbp]
	\begin{minipage}{1\linewidth}
		\centering
		\caption{Comparison of weakly-supervised semantic segmentation methods on COCO validation set in terms of mIoU (\%). All methods are based on the VGG-16 backbone.}
		\label{tab:comp_coco}
		\resizebox{\textwidth}{!}{%
			\begin{tabular}{l|c|c|c||l|c|c|c}
				\hline
				class           &SEC  &DSRG &Ours &class           &SEC  &DSRG &Ours \\ \hline 
				background      &74.3 &80.6 &73.9 &wine glass      &22.3 &24.0 &28.4 \\ \hline 
				person          &43.6 &-    &53.8 &cup             &17.9 &20.4 &29.3 \\ \hline 
				bicycle         &24.2 &30.4 &45.6 &fork            &1.8  &0.0  &14.5 \\ \hline 
				car             &15.9 &22.1 &35.5 &knife           &1.4  &5.0  &7.7  \\ \hline 
				motorcycle      &52.1 &54.2 &67.4 &spoon           &0.6  &0.5  &4.1  \\ \hline 
				airplane        &36.6 &45.2 &66.8 &bowl            &12.5 &18.8 &19.4 \\ \hline 
				bus             &37.7 &38.7 &66.0 &banana          &43.6 &46.4 &48.1 \\ \hline 
				train           &30.1 &33.2 &65.0 &apple           &23.6 &24.3 &32.1 \\ \hline 
				truck           &24.1 &25.9 &44.3 &sandwich        &22.8 &24.5 &40.6 \\ \hline 
				boat            &17.3 &20.6 &37.4 &orange          &44.3 &41.2 &43.2 \\ \hline 
				traffic light   &16.7 &16.2 &16.6 &broccoli        &36.8 &35.7 &34.2 \\ \hline 
				fire hydrant    &55.9 &60.4 &58.6 &carrot          &6.7  &15.3 &23.8 \\ \hline 
				stop sign       &48.4 &51.0 &47.1 &hot dog         &31.2 &24.9 &38.1 \\ \hline 
				parking meter   &25.2 &26.3 &53.6 &pizza           &50.9 &56.2 &62.5 \\ \hline 
				bench           &16.4 &22.3 &24.7 &donut           &32.8 &34.2 &49.2 \\ \hline 
				bird            &34.7 &41.5 &54.5 &cake            &12.0 &6.9  &40.3 \\ \hline 
				cat             &57.2 &62.2 &73.4 &chair           &7.8  &9.7  &14.7 \\ \hline 
				dog             &45.2 &55.6 &63.1 &couch           &5.6  &17.7 &22.8 \\ \hline 
				horse           &34.4 &42.3 &64.9 &potted plant    &6.2  &14.3 &11.1 \\ \hline 
				sheep           &40.3 &47.1 &60.8 &bed             &23.4 &32.4 &35.8 \\ \hline 
				cow             &41.4 &49.3 &63.2 &dining table    &0.0  &3.8  &6.4  \\ \hline 
				elephant        &62.9 &67.1 &81.3 &toilet          &38.5 &43.6 &48.9 \\ \hline 
				bear            &59.1 &62.6 &77.4 &tv              &19.2 &25.3 &33.5 \\ \hline 
				zebra           &59.8 &63.2 &66.8 &laptop          &20.1 &21.1 &36.8 \\ \hline 
				giraffe         &48.8 &54.3 &61.3 &mouse           &3.5  &0.9  &21.9 \\ \hline 
				backpack        &0.3  &0.2  &9.1  &remote          &17.5 &20.6 &22.1 \\ \hline 
				umbrella        &26.0 &35.3 &42.5 &keyboard        &12.5 &12.3 &42.2 \\ \hline 
				handbag         &0.5  &0.7  &2.9  &cell phone      &32.1 &33.0 &30.8 \\ \hline 
				tie             &6.5  &7.0  &3.7  &microwave       &8.2  &11.2 &24.7 \\ \hline 
				suitcase        &16.7 &23.4 &36.7 &oven            &13.7 &12.4 &24.8 \\ \hline 
				frisbee         &12.3 &13.0 &26.1 &toaster         &0.0  &0.0  &0.0  \\ \hline 
				skis            &1.6  &1.5  &4.2  &sink            &10.8 &17.8 &18.2 \\ \hline 
				snowboard       &5.3  &16.3 &14.3 &refrigerator    &4.0  &15.5 &24.3 \\ \hline 
				sports ball     &7.9  &9.8  &9.0  &book            &0.4  &12.3 &24.3 \\ \hline 
				kite            &9.1  &17.4 &14.7 &clock           &17.8 &20.7 &17.6 \\ \hline 
				baseball bat    &1.0  &4.8  &2.7  &vase            &18.4 &23.9 &11.3 \\ \hline 
				baseball glove  &0.6  &1.2  &0.2  &scissors        &16.5 &17.3 &18.0 \\ \hline 
				skateboard      &7.1  &14.4 &16.2 &teddy bear      &47.0 &46.3 &45.4 \\ \hline 
				surfboard       &7.7  &13.5 &21.8 &hair drier      &0.0  &0.0  &0.0  \\ \hline 
				tennis racket   &9.1  &6.8  &11.8 &toothbrush      &2.8  &4.5  &7.1  \\ \hline 
				bottle          &13.2 &22.3 &24.6 &\textbf{mean IoU} &\textbf{22.4} &\textbf{26.0} &\textbf{33.6} \\ \hline 
				
			\end{tabular}
		}
	\end{minipage}
\end{table}

\subsubsection{Semi-supervised Semantic Segmentation}
We compare our approach with other state-of-the-art semi-supervised semantic segmentation methods on PASCAL VOC 2012 dataset under the same setting, where 1.4K images annotated with pixel-level labels and 9K images annotated with image-level tags are available. The comparison results are reported in Table~\ref{tab:comp_semi}, together with the results obtained by the fully supervised Deeplab-ASPP~\cite{chen2017deeplab} segmentation network. Table~\ref{tab:comp_semi} shows that our approach not only outperforms all previous methods but also reaches 95.3\% of the performance under full supervision. It needs to be mentioned that most previous methods use the 1.4K pixel-level labels for training the semantic segmentation model only, while our approach can easily adopt them to facilitate the training of SGAN and improve the quality of dense seeds for the 9K weakly annotated images, leading to better performance.

\begin{table}[htbp]
	\begin{minipage}{1\linewidth}
		\centering
		\caption{Comparison of semi-supervised semantic segmentation methods on PASCAL VOC 2012 validation set in terms of mIoU. All methods are based on the VGG-16 backbone.} 
		\label{tab:comp_semi}
		\begin{tabular}{l|c|c|c}
			\hline
			Methods   							&Publication	&Training Set   		&mIoU \\ \hline
			Deeplab~\cite{chen2017deeplab}  	&TPAMI'17		&1.4K strong 			&62.5 \\ 
			& 			 	&10.6K strong 			&70.3 \\ \hline			
			WSSL~\cite{papandreou2015weakly} 	&ICCV'15		&1.4K strong + 9K weak 	&64.6 \\
			GAIN~\cite{li2018tell} 				&CVPR'18		&1.4K strong + 9K weak 	&60.5 \\
			MDC~\cite{wei2018revisiting} 		&CVPR'18		&1.4K strong + 9K weak 	&65.7 \\
			DSRG~\cite{huang2018weakly} 		&CVPR'18		&1.4K strong + 9K weak 	&64.3 \\
			FickleNet~\cite{lee2019ficklenet} 	&CVPR'19		&1.4K strong + 9K weak 	&65.8 \\
			SGAN(Ours)  						&-				&1.4K strong + 9K weak 	&\textbf{67.0} \\ \hline
		\end{tabular}
	\end{minipage}
\end{table}


\begin{figure}[htbp]
	\centering
	\includegraphics[width=.48\textwidth]{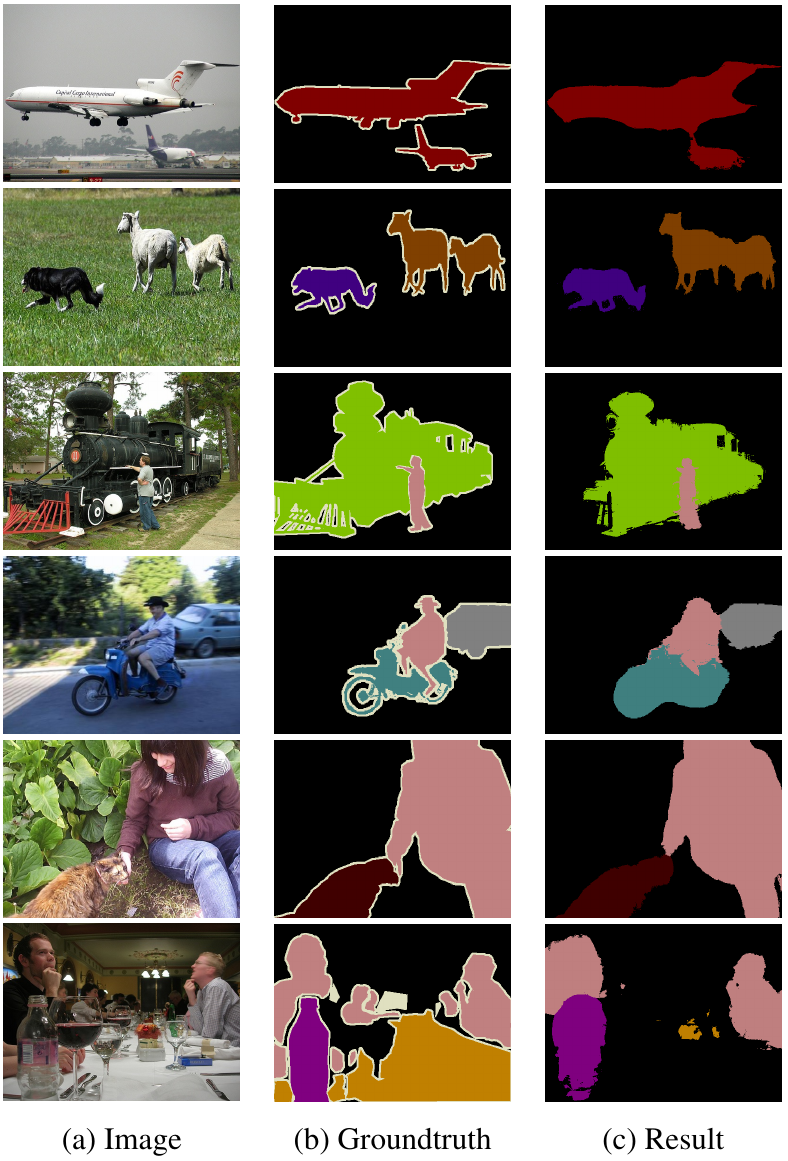}
	\caption{Examples of segmentation results on PASCAL VOC 2012 validation set obtained by the proposed approach.}
	\label{fig:seg_res}
\end{figure}

\begin{table*}[htbp]
	\caption{Comparison of the proposed model under different settings on VOC 2012 val set in terms of mIoU (\%). }
	\label{tab:ablation}
	\resizebox{\linewidth}{!}{
		\begin{tabular}{l|c|*{21}{p{0.25cm}}}
			\hline
			Method &mIoU  &\myrot{bkg}  &\myrot{aero}  &\myrot{bike} &\myrot{bird} &\myrot{boat} &\myrot{bottle} &\myrot{bus}  &\myrot{car}  &\myrot{cat}  &\myrot{chair} &\myrot{cow}  &\myrot{table} &\myrot{dog}  &\myrot{horse} &\myrot{motor} &\myrot{person} &\myrot{plant} &\myrot{sheep} &\myrot{sofa} &\myrot{train} &\myrot{tv} \\ \hline
			Baseline       &55.0  &86.5 &68.1  &29.8 &71.8 &56.2 &56.3   &47.6 &69.7 &75.4 &18.6  &60.6 &18.3  &62.6 &62.1  &67.1  &59.3   &34.4  &69.7  &27.3 &58.4  &55.0 \\
			SGAN-SAL-SEED  &45.8  &78.7 &51.4  &22.1 &23.5 &21.4 &62.5   &73.8 &60.2 &80.6 &6.6   &58.1 &4.3   &69.5 &45.8  &65.3  &66.1   &31.4  &35.4  &23.7 &48.7  &35.3 \\
			SGAN-SEED      &62.4  &89.5 &75.4  &31.0 &75.1 &60.0 &66.3   &68.3 &73.8 &82.3 &23.0  &74.8 &25.1  &76.2 &69.0  &69.1  &72.8   &40.3  &71.5  &32.8 &73.2  &60.6 \\
			SGAN$^{CLS}$  &63.7  &89.6 &75.0  &31.8 &73.1 &61.1 &67.4   &79.1 &75.4 &82.3 &26.3  &75.0 &28.5  &75.7 &67.8  &70.1  &73.1   &45.7  &72.5  &35.6 &73.2  &58.6 \\
			SGAN$^{SEG}$   &63.4  &89.9 &80.6  &34.6 &76.1 &60.4 &70.0   &75.7 &72.0 &82.8 &20.0  &76.7 &16.8  &76.8 &71.2  &70.1  &71.8   &45.1  &73.5  &35.2 &75.2  &57.2 \\
			SGAN           &\textbf{64.2}  &89.9 &77.7  &33.7 &75.3 &61.7 &68.5   &76.8 &76.3 &81.7 &28.7  &75.8 &27.4  &75.6 &70.0  &70.6  &73.3   &41.8  &73.5  &35.6 &74.9  &59.8  \\ \hline
		\end{tabular}
	}
\end{table*}

\begin{figure*}[t]
	\centering
	\includegraphics[width=.95\textwidth]{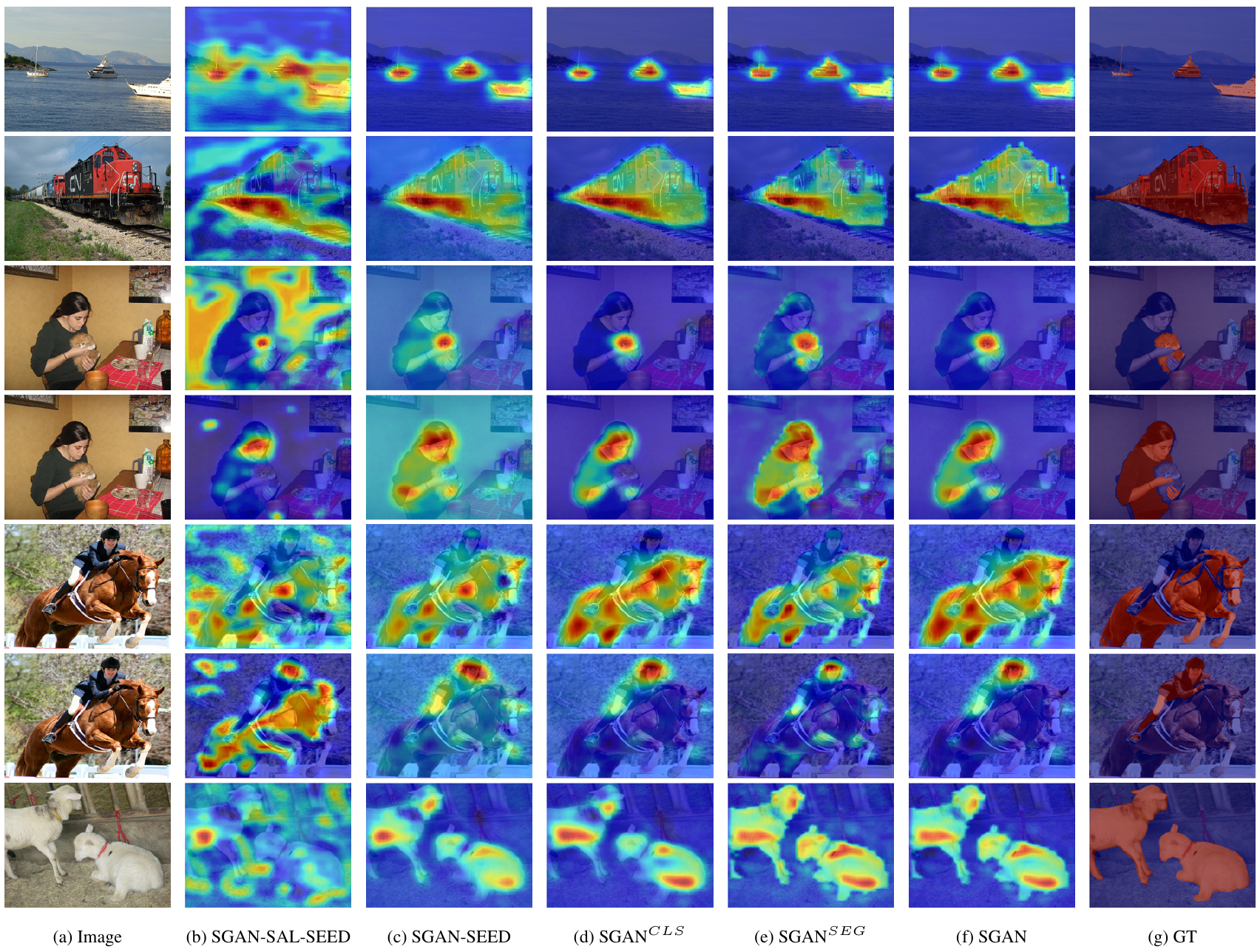}
	\caption{Visualization of the class attention maps learned by different variants of our SGAN.}
	\label{fig:cams}
\end{figure*}

\subsection{Qualitative Results}
Figure~\ref{fig:seg_res} shows qualitative segmentation results obtained by the proposed approach on PASCAL VOC dataset. As we can see, our approach produces accurate segmentation results and recovers fine details of object boundaries for images containing scale variation, multiple objects, and complex background. A typical failure case is also presented in the last row, in which the dining table is indistinguishable from the background and thus misidentified as background. 

\subsection{Ablation Studies}
\label{sec:ablation}
\subsubsection{Effectiveness of The Components in SGAN}
To investigate the effectiveness of each component in the saliency guided self-attention network, we conduct a series of experiments in different settings while keeping the VGG-16-based segmentation model the same throughout all these experiments. Particularly, the following configurations are investigated: (1) the full model, which is referred to as SGAN; (2) the full model but only using the CAMs from the image classification branch, which is denoted as SGAN$^{CLS}$;  (3) the full model but only using the CAMs from the seed segmentation branch, which is denoted as SGAN$^{SEG}$; (4) the model without the segmentation branch and the seed loss, which is denoted as SGAN-SEED; (5) the model without the segmentation branch, the seed loss, and the saliency guidance, which in essence is directly integrating the self-attention mechanism into the modified VGG-16 classification network. We denote this variant as SGAN-SAL-SEED; and (6) the baseline model without our proposed saliency guided self-attention module, which is actually the modified VGG-16 classification network.

The comparison results are listed in Table~\ref{tab:ablation}, from which we make the following observations: 
(1) The SGAN-SAL-SEED model that applies the self-attention mechanism directly to a weakly-supervised network degrades the segmentation performance, especially for the categories that are always co-occurrent with the same background, for instance, 'airplane' with 'sky', 'boat' with 'water', 'horse' with 'grass', etc. In such cases, SGAN-SAL-SEED tends to propagate attentions from foreground objects to the concurrent background and generate inaccurate seeds. (2) The SGAN-SEED model that uses the proposed saliency guided self-attention module outperforms the baseline model over all categories. (3) After integrating class-specific attention cues, both SGAN$^{CLS}$ and SGAN$^{SEG}$ produce high quality seeds and achieve high segmentation performance. The ensemble of CAMs from two branches, that is SGAN, boosts the performance further. 
(4) Compared to the baseline model, our full model boosts the performance significantly for the categories containing objects in large size, like 'bus' (+29.2\%) and 'train' (+16.5\%), and the categories with large scale variation such as 'person' (+14\%). For these categories, the initial localization cues are usually too sparse to delineate the integral object extent. Our model can effectively propagate attentions from small discriminative parts to non-discriminative regions of objects and generate more complete object seeds, leading to much better segmentation performance. 

In order to understand these models more intuitively, we present the class activation maps generated by each variant in Figure~\ref{fig:cams}.  From this figure we observe that SGAN-SAL-SEED tends to diffuse class-specific attentions to backgrounds and other categories. SGAN-SEED can greatly reduce the diffusion to backgrounds but it cannot prevent the miss-spread among foreground categories. SGAN$^{CLS}$, SGAN$^{SEG}$, and SGAN can constrain attentions mostly within the regions of the right class. Moreover, the CAMs obtained by SGAN$^{CLS}$ and SGAN$^{SEG}$ are complementary to each other in some cases such as the 'train' image and the 'sheep' image.  


\subsubsection{Evaluation of Seeds' Quality}
High-quality seeds are obtained by thresholding the class activation maps presented above. Here, we also adopt precision, recall, and the F-measure score to evaluate the quality of seeds produced by different variants of our SGAN. The F-measure is defined as the weighted harmonic mean of the precision and recall:
\begin{equation}
	F_\beta = \frac{(1 + \beta^2) \cdot Precision \cdot Recall}{\beta^2 \cdot Precision + Recall},
\end{equation}
where $\beta^2$ is empirically set to be 0.4 to emphasize the importance of precision. 

Table~\ref{tab:seed_quality} reports the evaluation results. From it we get the following observations: (1) The SGAN-SAL-SEED model that applies the self-attention mechanism directly under weakly-supervised settings degrades the seed's precision drastically and thus leading to poor segmentation performance. (2) The full model, SGAN, enhances the recall of seeds by a large margin while maintains the precision level, indicating that it can produce dense and accurate seeds. (3) The F-measure score shows a strong correlation with the final segmentation performance (mIoU). The higher a F-measure is, the better a variant SGAN model can perform. 


\begin{table}[htp]
	\caption{The quality of seeds generated by different variants of our SGAN model.}
	\label{tab:seed_quality}
	\centering
	\begin{tabular}{l|c|c|c|c}
		\hline
		Method			&Precision		&Recall	 &F\_score	&mIoU					\\ \hline
		Baseline				&75.5			&28.4		&61.5		&55.0			\\
		SGAN-SAL-SEED		&31.7			&47.6		&33.2		&45.8			  	\\
		SGAN-SEED				&76.1			&48.6			&70.5		&62.4	  	\\
		SGAN$^{CLS}$			&74.4			&60.0		&72.0		&63.7			\\
		SGAN$^{SEG}$		&73.8			&57.1		&70.9		&63.4			\\
		SGAN				&76.4			&57.4		&73.0		&64.2				\\ \hline
	\end{tabular}
\end{table}

\subsubsection{Visualization of Context Attention Maps}
\begin{figure}[t]
	\centering
	\includegraphics[width=.45\textwidth]{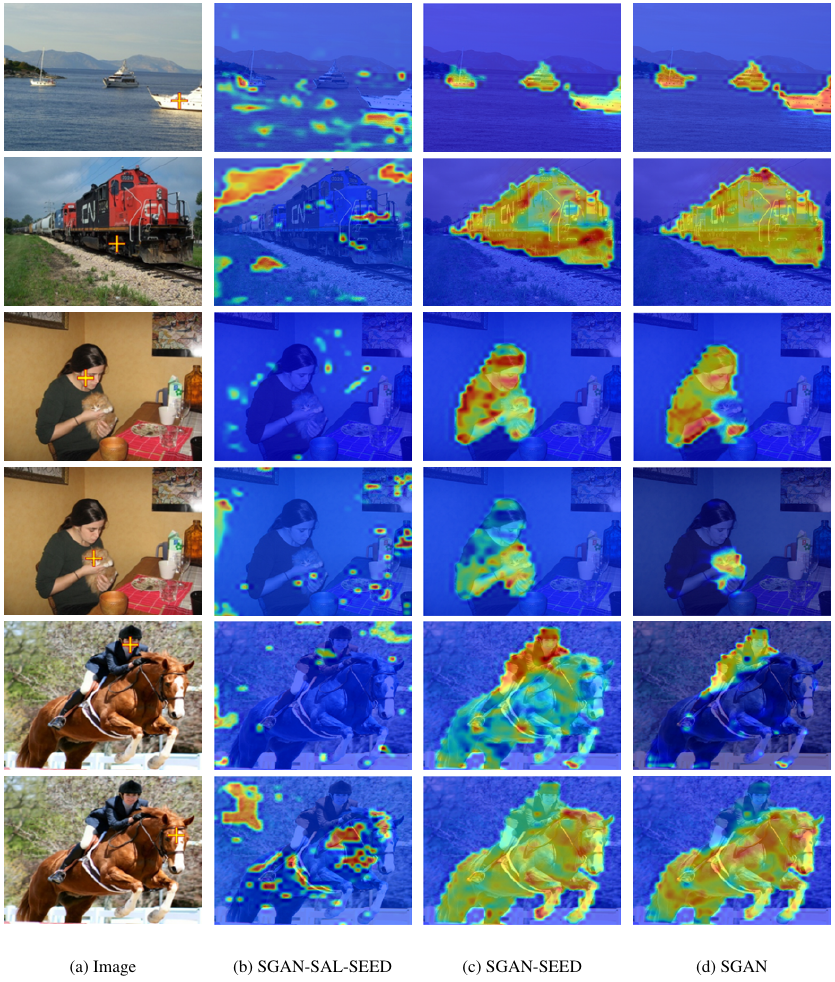}
	\caption{Visualization of the context attention maps learned by different variants of our SGAN.}
	\label{fig:self}
\end{figure}

To better understand how the self-attention mechanism behaves in our models, Figure~\ref{fig:self} visualizes the context attention maps learned by different variant of SGAN. Specifically, we select one discriminative pixel in each image and mark it by a yellow '+'. The attentions propagated from the selected pixel to all other pixels are indicated in the corresponding column of the learned context attention map. We reshape the column into the image size and overlay it on the color image for visualization. As shown in Figure~\ref{fig:self}, simply integrating the self-attention mechanism in a weakly-supervised network tends to mess up the attentions. Saliency priors can prohibit from spreading the attentions to the background. By further integrating the class-specific attention cues, our full model can restrict the attentions propagated mostly to the pixels belonging to the same category with the selected pixel. These maps help us to interpret the CAMs presented in Figure~\ref{fig:cams}.

\subsubsection{Influence of The Weighting Factor $\lambda$}
The weighting factor $\lambda$ in the total loss (in Equation (\ref{eq:sganloss})) of SGAN determines the impact of the seed loss. Without the seed loss, no class-specific attention cues are included and our SGAN cannot handle the problem of mis-spreading attentions among foreground categories. Whereas, putting too much weight on this term may cause inefficient training due to the sparsity of the seeds. Therefore we carry out a set of experiments to check the influence of $\lambda$ and report the results in Table~\ref{tab:lambda}. It shows that $\lambda=0.15$ leads to the best performance. Therefore, we empirically set this value throughout all other experiments. 


\begin{table}[htbp]
	\caption{Influence of the weighting factor $\lambda$ to the segmentation performance on PASCAL VOC 2012 validation set.}
	\label{tab:lambda}
	{\small
		\begin{tabular}{c|ccccccc}
			\hline
			$\lambda$ &0      &0.05    &0.1    &0.15   			 &0.2    &0.25    &0.3 \\ \hline
			mIoU (\%) &62.4   &63.9    &64.0  &\textbf{64.2}    &64.1  &64.0   &63.6 \\ \hline
		\end{tabular}
	}
\end{table}

\subsubsection{Influence of Saliency Detectors}
As marked out in Table~\ref{tab:comp}, it is quite common for WSSS methods to take saliency as additional guidance. The reason is that it can provide rough localization of foreground objects and therefore help the propagation of CAMs. In our work, we adopt it to prevent object's discriminative information from spreading to unexpected background regions. 

Various saliency detectors, such as DHSNet~\cite{Liu2016saliency}, DSS~\cite{hou2017deeply}, and S-Net~\cite{xiao2018deep}, have been adopted in recent WSSS methods~\cite{chaudhry2017discovering,hou2018self,wei2018revisiting} to produce saliency masks. In our work, saliency masks are generated by S-Net~\cite{xiao2018deep}. We also conduct experiments using DHSNet~\cite{Liu2016saliency} and DSS~\cite{hou2017deeply}. Table~\ref{tab:sal} shows that SGAN is not so sensitive to saliency detectors, because these saliency detection methods perform comparable well on outdoor scenarios but all have difficulties in handling indoor scenes. In addition, we admit that the errors in saliency masks may have negative effects for the WSSS task. But our proposed method is still performing better than the recent WSSS method SSNet~\cite{Zeng_2019_ICCV} that jointly learns saliency and segmentation tasks, as shown in Table~\ref{tab:comp}. 

\begin{table}[htp]
	\caption{Influence of saliency detectors to the segmentation performance on PASCAL VOC 2012 validation set.}
	\label{tab:sal}
	\centering
	\begin{tabular}{c|c|c|c}
		\hline
		saliency mask &S-Net~\cite{xiao2018deep} &DSS~\cite{hou2017deeply}  &DHSNet~\cite{Liu2016saliency} \\ \hline
		mIoU (\%)     &64.2  &64.0 &63.7 \\ \hline
	\end{tabular}
	
\end{table}

\section{Conclusion}
In this paper, we have presented a saliency guided self-attention network to address the semantic segmentation problem supervised by image-level labels only. To generate dense and accurate object seeds, we introduced the self-attention mechanism into the weakly-supervised scenario and utilized both class-agnostic saliency maps and class-specific attention cues to enable the mechanism work effectively. Extensive experiments on PASCAL VOC 2012 dataset show that the proposed method outperforms the baseline model with a large margin and performs better than all other state-of-the-art methods.

\bibliographystyle{IEEEtran}
\bibliography{SGAN}

\end{document}